# Learning STRIPS Operators from Noisy and Incomplete Observations

**Kira Mourão**
School of Informatics
University of Edinburgh
Edinburgh, EH8 9AB, UK
kmourao@inf.ed.ac.uk

**Luke Zettlemoyer**
Computer Science & Engineering
University of Washington
Seattle, WA98195
lsz@cs.washington.edu

**Ronald P. A. Petrick**
School of Informatics
University of Edinburgh
Edinburgh, EH8 9AB, UK
rpetrick@inf.ed.ac.uk

**Mark Steedman**
School of Informatics
University of Edinburgh
Edinburgh, EH8 9AB, UK
steedman@inf.ed.ac.uk

## Abstract

Agents learning to act autonomously in real-world domains must acquire a model of the dynamics of the domain in which they operate. Learning domain dynamics can be challenging, especially where an agent only has partial access to the world state, and/or noisy external sensors. Even in standard STRIPS domains, existing approaches cannot learn from noisy, incomplete observations typical of real-world domains. We propose a method which learns STRIPS action models in such domains, by decomposing the problem into first learning a transition function between states in the form of a set of classifiers, and then deriving explicit STRIPS rules from the classifiers' parameters. We evaluate our approach on simulated standard planning domains from the International Planning Competition, and show that it learns useful domain descriptions from noisy, incomplete observations.

## 1 INTRODUCTION

Developing agents with the ability to act autonomously in the world is a major goal of artificial intelligence. One important aspect of this development is the acquisition of domain models to support planning and decision-making: to operate effectively in the world, an agent must be able to accurately predict when its actions will succeed, and what effects its actions will have. Only when a reliable action model is acquired can the agent usefully combine sequences of actions into plans, in order to achieve wider goals. However, learning domain dynamics can be a challenging problem: agents' observations may be noisy, or incomplete; actions may be non-deterministic; the world may be noisy or contain many irrelevant objects and relations.

In this paper we consider the problem of acquiring explicit domain models from the raw experiences of an agent exploring the world, where the agent's observations are incomplete, and observations and actions are subject to noise.

The domains we consider are relational STRIPS (Fikes and Nilsson, 1971) domains, although our approach has the potential to be extended to more expressive domains. Given the autonomous learning setting, we assume only a weak domain model where the agent knows how to identify objects, has acquired predicates to describe object attributes and relations, and knows what types of actions it may perform, but not the appropriate contexts for the actions, or their effects. Experience in the world is then developed through observing changes to object attributes and relations when motor-babbling with primitive actions.

Other approaches to learning STRIPS operators do not handle both noisy and incomplete observations (see Section 3). We develop a two-stage approach to the problem which decouples the requirement to tolerate noisy, incomplete observations from the requirement to learn compact STRIPS operators. In the first stage we learn action models by constructing a set of kernel classifiers which tolerate noise and partial observability, but whose action models are implicit in the learnt parameters of the classifiers, similar to the work of Mourão *et al.* (2009, 2010). However, we additionally use the method to learn preconditions as well as effects, as suggested but not explored in earlier work. Also, we evaluate additional kernels for the learning problem and select a better performing alternative. The initial action model learnt in this first stage acts as a noise-free, fully observable source of observations from which to extract explicit action rules. In the second stage we devise a novel method to derive explicit STRIPS operators from the model implicit in the kernel classifiers. In experiments the resulting rules perform as well as the original classifiers, while providing a compact representation of the action models suitable for use in automated planning systems.

## 2 THE LEARNING PROBLEM

A *domain* is a tuple $\mathcal{D} = \langle \mathcal{O}, \mathcal{P}, \mathcal{A} \rangle$, where $\mathcal{O}$ is a finite set of world objects, $\mathcal{P}$ is a finite set of predicate (relation) symbols, and $\mathcal{A}$ is a finite set of actions. Each predicate and action also has an associated arity. A *fluent expression* is a statement of the form $p(c_1, c_2, ..., c_n)$, where $p \in \mathcal{P}$,

$n$ is the arity of $p$, and each $c_i \in \mathcal{O}$. A *state* is any set of fluent expressions, and $\mathcal{S}$ is the set of all possible states. Since state observations may be incomplete we assume an open world where unobserved fluents are considered to be unknown. For any state $s \in \mathcal{S}$, a fluent expression $\phi$ is true at $s$ iff $\phi \in s$. The negation of a fluent expression, $\neg \phi$, is true at $s$ (also, $\phi$ is false at $s$) iff $\neg \phi \in s$. If $\phi \in s$ then $\neg \phi \notin s$. Any (legal) fluent expression not in $s$ is unobserved.

Each action $a \in \mathcal{A}$ is defined by a set of *preconditions*, $Pre_a$, and a set of *effects*, $Eff_a$. In STRIPS domains $Pre_a$ can be any set of fluent expressions and $Eff_a$ can be any set of fluent expressions and negated fluent expressions. Action preconditions and effects can also be parameterised. An action with all of its parameters replaced with objects from $\mathcal{O}$ is said to be an *action instance*. Objects mentioned in the preconditions or the effects must be listed in the action parameters (the *STRIPS scope assumption* (SSA)).

The task of the learning mechanism is to learn the preconditions and effects $Pre_a$ and $Eff_a$ for each $a \in \mathcal{A}$, from data generated by an agent performing a sequence of randomly selected actions in the world and observing the resulting states. The sequence of states and action instances is denoted by $s_0, a_1, s_1, ..., a_n, s_n$ where $s_i \in \mathcal{S}$ and $a_i$ is an instance of some $a \in \mathcal{A}$. Our data consists of *observations* of the sequence of states and action instances $s'_0, a_1, s'_1, ..., a_n, s'_n$, where state observations may be noisy (some $\phi \in s_i$ may be observed as $\neg \phi \in s'_i$) or incomplete (some $\phi \in s_i$ are not in $s'_i$). Action failures are allowed: the agent may attempt to perform actions whose preconditions are unsatisfied. In these cases the world state does not change, but the observed state may still be noisy or incomplete. To make accurate predictions in domains where action failures are permitted, the learning mechanism must learn both preconditions and effects of actions.

E.g. in BlocksWorld, where an agent stacks and unstacks blocks, consider a state with blocks `B1` and `B2` on the table:

```
(AND armempty (ontable B1) (ontable B2) (clear B1)
 (clear B2) (NOT (on B1 B2)) (NOT (on B2 B1))
 (NOT (holding B1)) (NOT (holding B2))).
```

A corresponding noisy, incomplete observation is:

```
(AND armempty (ontable B2) (holding B1) (clear B1)
 (NOT (on B1 B2)) (NOT (holding B2))).
```

It is noisy as `(holding B1)` is incorrect, and incomplete as `(ontable B1)`, `(clear B2)` and `(NOT(on B2 B1))` are missing. A sequence of noisy, incomplete observations of states and actions could be as follows:

$s'_0$: `(AND armempty (ontable B2) (holding B1)`
`    (clear B1) (NOT (on B1 B2)) (NOT (holding B2)))`
$a_1$: `(pickup B1)`
$s'_1$: `(AND (NOT (clear B1)) (holding B1) (NOT (on B1 B2))`
`    (NOT (ontable B2)) (NOT (on B2 B1)))`
$a_2$: `(stack B1 B2)` (1)
$s'_2$: `(AND armempty (clear B1) (clear B2)`
`    (ontable B2) (NOT (holding B1)))`
$a_3$: `(stack B2 B1)`
$s'_3$: `(AND armempty (clear B1) (NOT (clear B2))`
`    (NOT (on B2 B1)) (NOT (on B2))).`

Taking a sequence of such inputs, we learn action descriptions for each action in the domain. For example, the `stack` action, which moves a block from the gripper on to another block, would be represented as:

```
(:action stack
 :parameters (?ob ?underob)
 :precondition (and (clear ?underob) (holding ?ob))
 :effect (and (arm-empty) (clear ?ob) (on ?ob ?underob)
        (not (clear ?underob)) (not (holding ?ob)))).
```

## 3 RELATED WORK

Previous work fulfils some but not all of the requirements for learning models in the setting we consider. In autonomous robotics, various techniques exist to learn preconditions and effects of actions in noisy, partially observable worlds (Doğar *et al.*, 2007; Metta and Fitzpatrick, 2003; Modayil and Kuipers, 2008). However, none of these approaches learn relational models. Conversely, much previous work on learning relational action models relies on the provision of prior knowledge of the action model. Strategies include seeding initial models with approximate planning operators (Gil, 1994), using known successful plans (Wang, 1995), excluding action failures (Amir and Chang, 2008), or the presence of a teacher (Benson, 1996). Such knowledge is unlikely to be available to an autonomous agent learning the dynamics of its domain. Additionally, only a few approaches are capable of learning under either partial observability (Amir and Chang, 2008; Yang *et al.*, 2007; Zhuo *et al.*, 2010), noise in any form (Pasula *et al.*, 2007; Rodrigues *et al.*, 2010), or both (Halbritter and Geibel, 2007; Mourão *et al.*, 2010). Approaches which handle both do so by generating implicit action models which must be used as a black-box to make predictions of state changes, and do not generate symbolic action representations.

We extract rules from classifiers based on the intuition that more discriminative features will contribute more to the classifier's objective function. This is similar to the insight underlying feature selection methods of the type which rank each feature according to the sensitivity of the classifier's objective function to the removal of the feature (Kohavi and John, 1997; Guyon *et al.*, 2002). Our approach differs in that features are selected separately for individual examples, rather than once across the entire training set, and we define a stopping criterion which identifies when the set of selected features can no longer form a rule.

Our work also has links with earlier work in version spaces (Mitchell, 1982) and the associated greedy search, which underlie many other approaches to rule learning (e.g. Amir and Chang, 2008; Pasula *et al.*, 2007). Our rule search benefits from extra information to guide the search, in the form of the weights associated with each hypothesis by the previously trained statistical classifiers, which can be highly robust to noise and incomplete observations.

# 4 LEARNING IMPLICIT MODELS

The basis of our approach is the division of the learning problem into two parts: initially a classification method is used to learn to predict effects of actions, then STRIPS rules are derived from the resulting action representations. We define the *implicit action model* to be the model of the domain implicit in the learnt parameters of the classifiers, and the *explicit action model* to be the domain model described by STRIPS rules. In learning an implicit action model, our approach follows earlier work (Croonenborghs *et al.*, 2007; Halbritter and Geibel, 2007; Mourão *et al.*, 2009, 2010) which encodes the learning problem in terms of the inputs and outputs of a set of classifiers. However, none of these methods generate explicit action models.

Following the approach of Mourão *et al.* (2009, 2010), we encode the state descriptions with a fixed dimension vector representation by only considering objects which are mentioned in the action parameter list. By the SSA this is sufficient to learn STRIPS rules. For an action instance with arguments $o_1, o_2, ..., o_n$ we therefore restrict the state description to all possible fluents whose arguments are in $\{o_1, o_2, ..., o_n\}$. Also we schematise descriptions by replacing the $i$-th action parameter with the label $arg_i$ whenever it appears in any fluent. Thus all state descriptions are now written in terms of the action parameters and not in terms of specific objects. The SSA fixes a small number of objects to consider for an action, as well as their roles, which allows relational state descriptions to be encoded in a vector, as each possible fluent in a state maps to exactly one possible fluent in any other state. We encode state descriptions as vectors where each bit in the vector corresponds to each possible fluent which could exist in the schematised state description. The value of a bit is 1 if the fluent is true, $-1$ if false, and a wildcard value * if unobserved.

In our previous BlocksWorld example (1), the state descriptions for $s'_0$ and $s'_1$ in the context of the pickup(B1) action $a_1$ would include only B1, and are schematised to form prior and successor states $s_{prior}$ = (AND armempty (holding arg$_1$) (clear arg$_1$)) and $s_{succ}$ = (AND (holding arg$_1$) (NOT(clear arg$_1$))) respectively, to give $\langle s_{prior}, \text{pickup}(arg_1), s_{succ} \rangle$ as a training example. Conversely, the descriptions for $s'_1$ and $s'_2$ in the context of the stack(B1 B2) action $a_2$ contain both B1 and B2 as both are action parameters, giving $s_{prior}$ = (AND (holding arg$_1$) (NOT(on arg$_1$ arg$_2$)) (NOT(on arg$_2$ arg$_1$)) (NOT(clear arg$_1$)) (NOT(ontable arg$_2$))) and $s_{succ}$ = (AND armempty (clear arg$_1$) (clear arg$_2$) (ontable arg$_2$) (NOT(holding arg$_1$))).

Encoding the state descriptions as vectors, where the bits correspond to armempty followed by clear, ontable, holding and on with first argument arg$_1$, and then the same predicates with first argument arg$_2$, we obtain $v^{prior} = \langle 1,1,*,1,*,*,*,*,* \rangle$ and $v^{succ} = \langle *,-1,*,1,*,*,*,*,* \rangle$ for $a_1$ and $v^{prior} = \langle *,-1,*,1,-1,*,-1,*,-1 \rangle$ and $v^{succ} = \langle 1,1,*,-1,*,1,1,*,* \rangle$ for $a_2$.

Given the vectorised state descriptions, a changes vector $v^{diff}$ is derived for each training example, where the $i$-th bit $v_i^{diff}$ is defined as follows:

$$v_i^{diff} = \begin{cases} 0, & \text{if } v_i^{prior} = v_i^{succ} \text{ and } v_i^{succ}, v_i^{prior} \in \{-1,1\} \\ 1, & \text{if } v_i^{prior} \neq v_i^{succ} \text{ and } v_i^{succ}, v_i^{prior} \in \{-1,1\} \\ *, & \text{otherwise.} \end{cases}$$

A set of classifiers now learns the implicit action model. Each classifier $C_{a,i}$ corresponds to a particular action $a$ in the domain and predicts the $i$-th bit of the changes vector. Thus if $\langle s_{prior}, a, s_{succ} \rangle$ is a training example for $C_{a,i}$ then the input to the classifier is $v^{prior}$ with target value $v_i^{diff}$.

Voted perceptron classifiers equipped with a kernel function have previously been applied to the problem of learning action models (Mourão *et al.*, 2009, 2010) as they are computationally efficient and known to tolerate noise (Khardon and Wachman, 2007). We take a similar approach. Mourão *et al.* used voted perceptrons combined with a DNF kernel, $K(x,y) = 2^{same(x,y)}$, where $same(x,y)$ is the number of bits with equal values in $x$ and $y$ (Sadohara, 2001; Khardon and Servedio, 2005). In the $same(x,y)$ calculation, bits with unobserved values are excluded. The features of the DNF kernel are all possible conjunctions of fluents, seemingly ideal for learning action preconditions which are arbitrary conjunctions of fluents. However, DNF is not PAC-learnable by a perceptron using the DNF kernel, as examples exist on which it can make exponentially many mistakes (Khardon *et al.*, 2005). We therefore consider as an alternative the k-DNF kernel, whose features are all possible conjunctions of fluents of length $\leq k$ for some fixed $k$: $K(x,y) = \sum_{l=0}^{k} \binom{same(x,y)}{l}$ (Khardon and Servedio, 2005). Preconditions with more than $k$ fluents are still possible since the voted perceptron supports hypotheses which are conjunctions of features.

# 5 RULE EXTRACTION

Once the classifiers have been trained, the first step in deriving explicit action rules is to extract individual *per-effect* rules to predict each fluent in isolation. We will look at how to combine the rules to extract postconditions in Section 6.

The rule extraction process takes as input a classifier $C_{a,i}$ and returns a set of preconditions (as vectors) which predict that the fluent corresponding to bit $i$ will change if action $a$ is performed. For example, in the BlocksWorld domain a set of preconditions extracted from the classifiers for stack is shown in Figure 2.

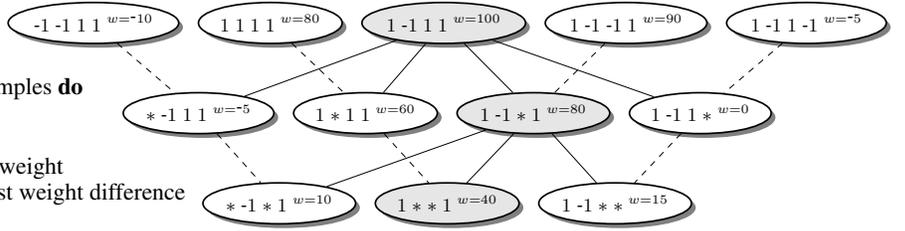

```
for v ∈ SV⁺ do
    child := v
    while child only covers +ve training examples do
        parent := child
        for each valued bit in parent do
            flip bit to its negation and calculate weight
        child := child whose parents have least weight difference
    rule_v := parent
```

(a) Rule Extraction Algorithm  (b) Example of Rule Extraction Process

Figure 1: Each node in (b) contains a vector corresponding to a possible precondition, and the weight $w$ assigned to the vector by the voted perceptron model. Each level of the lattice contains vectors with one fewer feature than the level above. Lines join parent and children nodes: solid lines link the candidate parent rule at one level with its children in the level below, and dashed lines link children to their alternative parent. Shaded nodes are the preconditions selected at each iteration through the lattice. The positive support vector "seed" is the vector $\langle 1\ \text{-}1\ 1\ 1\rangle$ with weight 100. Following the rule extraction algorithm in (a), the child whose parents have the least weight difference, the vector $\langle 1\ \text{-}1\ *\ 1\rangle$, is chosen as the next candidate rule. The process ends with the rule $\langle 1\ *\ *\ 1\rangle$ as both children have a negative counterexample in the training data (not shown).

```
(armempty) changes when:
[8]   (AND (NOT(armempty)) (NOT(ontable arg₁)))

(clear arg₁) changes when:
[14]  (AND (NOT(clear arg₁)) (holding arg₁)
           (NOT(on arg₁ arg₂)))
[12]  (AND (NOT(clear arg₁)) (NOT(ontable arg₁))
           (NOT(on arg₂ arg₁)))
[8]   (AND (clear arg₁) (ontable arg₁) (clear arg₂)
           (NOT(on arg₁ arg₂)) (NOT(on arg₂ arg₁)))

(ontable arg₁) changes when:
[6]   (AND (NOT(ontable arg₁)) (NOT(on arg₁ arg₂)))
[4]   (AND (NOT(armempty)) (ontable arg₁)
           (NOT(clear arg₂)))

(holding arg₁) changes when:
[15]  (AND (holding arg₁))

(on arg₁ arg₂) changes when:
[3]   (AND (NOT(on arg₁ arg₂)))

(clear arg₂) changes when:
[12]  (AND (clear arg₂) (ontable arg₂)
           (NOT(holding arg₂)))
[6]   (AND (NOT(armempty)) (NOT(clear arg₁)) (clear arg₂)
           (holding arg₁) (NOT(on arg₁ arg₂)))
[2]   (AND (NOT(clear arg₁)) (clear arg₂)
           (NOT(ontable arg₂)))
```

Figure 2: Per-effect rules generated for the BlocksWorld `stack` action from 1000 examples in a world with 5% noise and 25% observability. Weights are shown in square brackets. Fluents in bold are neither in, nor implied by, the true action specification. Many of these fluents will later be excluded by the rule combination process (Section 6).

Rules are extracted from a voted perceptron with kernel $K$ and support vectors $SV = SV^+ \cup SV^-$, where $SV^+$ ($SV^-$) is the set of support vectors whose *predicted* values are $1$ ($-1$). The positive support vectors are each instances of some rule learnt by the perceptron, and so are used to "seed" the search for rules. The extraction process aims to identify and remove all irrelevant bits in each support vector, using the voted perceptron's prediction calculation to determine which bits to remove.

The *weight* of any possible state description vector $\mathbf{x}$ is defined to be the value calculated by the voted perceptron's prediction calculation before thresholding (Freund and Schapire, 1999):

$$weight_e(\mathbf{x}) = \sum_{i=1}^{n} c_i\ sign \sum_{j=1}^{i} y_j \alpha_j K(\mathbf{x}_j, \mathbf{x})$$

where each $x_i$ is one of the $n$ support vectors, $y_i$ is the corresponding target value, $c_i$ and $\alpha_i$ are the parameters learnt by the classifier, and $e$ is the effect predicted by the classifier. The predicted value for $\mathbf{x}$ is 1 if $weight_e(\mathbf{x}) > 0$ and $-1$ otherwise. A *child* of vector $\mathbf{x}$ is any distinct vector obtained by replacing a single bit of $\mathbf{x}$ with the value *. Similarly, a *parent* of $\mathbf{x}$ is any vector obtained by replacing a *-valued bit with the value $1$ or $-1$.

The basic intuition behind the rule extraction process is that more discriminative features will contribute more to the weight of an example. Thus the rule extraction process operates by taking each positive support vector and repeatedly deleting the feature which contributes least to the weight until some stopping criterion is satisfied. This leaves the most discriminative features underlying the example, which can be used to form a precondition. An example of the process of extracting rules is shown in Figure 1(b), and an outline of the algorithm in Figure 1(a), as follows. Take each positive support vector $v$ in turn, and aim to find a conjunction $rule_v$ which covers $v$ and does not cover any negative training examples, but where every child of $rule_v$ covers at least one negative example. Construct $rule_v$ by a greedy algorithm which first takes $v$ as a candidate rule and then repeatedly creates a new candidate rule by choosing one bit to set to the $*$ value. The bit is chosen by considering the difference in weights between the current candidate $\mathbf{x} = \langle x_1, ..., x_i, ..., x_n \rangle$ and

each $\mathbf{x}_{\neg i} = \langle x_1, ..., \neg x_i, ..., x_n \rangle$, finding

$$\operatorname*{argmin}_{x_i \in \{x_1,...,x_i,...,x_n\}} (weight_e(\mathbf{x}) - weight_e(\mathbf{x}_{\neg i})).$$

Removing the resulting $x_i$ removes the least discriminative bit in the current candidate rule. At each step the new candidate rule is tested against the training examples. If it classifies a negative training example as positive, then the rule is too general and $rule_v$ is set to the previous candidate rule, otherwise the process repeats. The result is a set of rules for each action, predicting when a particular output bit changes. There may be many rules, up to one per positive support vector, each consisting of a set of preconditions which, if satisfied, predict the output bit will change.

## 6  RULE COMBINATION

The rule extraction process described above produces a set of rules for an action, such as for the BlocksWorld `stack` action shown in Figure 2. However, in STRIPS we expect a single rule for each action, consisting of a set of preconditions and a set of effects such as in the definition of the BlocksWorld `stack` action given in Section 2.

The rule combination process therefore builds a single STRIPS-like rule for each action, taking as input the set of rules produced by the rule extraction process: $\{(v_1, e_1), (v_2, e_2), \ldots, (v_r, e_r)\}$ where $v_i$ is the vector representing the $i$-th set of preconditions, and $e_i$ is the bit which changes when $v_i$ holds. Rule combination generates a rule $(v_{rule}, e_{rule})$ where the $j$-th element of $v_{rule}$, $v_{rule,j} \subseteq \bigcup_i v_{i,j}$ and $e_{rule} \subseteq \bigcup_i e_i$. Given the definitions in Section 4, we can directly convert the single state vector $v_{rule}$ and the set of effects $e_{rule}$ into a precondition and effect in STRIPS format.[1]

Without noise or partial observability, the combination process is a straightforward conjunction of all preconditions and all effects in the set of rules for an action, i.e., $\forall j \; v_{rule,j} = \bigcup_i v_{i,j}$ and $e_{rule} = \bigcup_i e_i$. However, when learning from noisy examples, unwanted additional fluents can be introduced to the per-effect rules via noisy support vectors. Similarly, incomplete training examples can mean some necessary fluents are missing from individual per-effect rules. In this section we describe an approach to identify and eliminate fluents introduced by noise while adding in fluents omitted due to partial observability.

To support the process of choosing between different potential rules which may contain noisy fluents, or omit necessary fluents, we introduce two filtering functions. $AcceptPrecons$ takes an existing precondition and effects $(v_{rule}, e_{rule})$ for an action $a$, and assesses whether a new

---
[1]Since the effects in $e_{rule}$ are changes it may be necessary to identify from what value the change is made, by referring to the rule from which the effect bit originated.

```
R := {(v_1, e_1), ..., (v_r, e_r)}
rule := (v_1, ∅)
locks = ∅
while R ≠ ∅ do
    next := highest weighted rule in R
    R := R \ {next}
    v_candidate = CombinePrecons(rule, next, locks)
    if v_candidate ≠ v_rule then
        v_candidate = SimplifyPrecons(rule, next, v_candidate)
        if AcceptPrecons(rule, v_candidate) then
            v_rule := v_candidate
        if AcceptEffect(rule, e_next) then
            e_rule := e_rule ∪ e_next
    e_rule = SimplifyEffects(rule)
```

Figure 3: Outline Rule Combination Algorithm

precondition $v_{new}$ predicts the effects of $a$ at least as well as the current precondition. $AcceptEffects$ takes an existing precondition and effects for an action $a$, and assesses whether the precondition predicts a new effect $e_{new}$ of $a$ at least as well as it predicts the current effects. We describe $AcceptPrecons$ and $AcceptEffects$ in detail in Section 6.2.

### 6.1  RULE COMBINATION OVERVIEW

With the filtering functions in place, we now describe how the rule combination process generates STRIPS rules by combining and refining the per-effect rules. Figure 3 gives an outline of the algorithm, described below.

For each action the process derives a rule $(v_{rule}, e_{rule})$ from the set of rules $R = \{(v_1, e_1), \ldots, (v_r, e_r)\}$ produced by rule extraction, ordered so that $weight_{e_i}(v_i) \geq weight_{e_j}(v_j)$ if $i < j$. The process first initialises $v_{rule}$ to the highest weighted precondition in $R$ and sets $e_{rule} = \varnothing$. The rule is then refined by combining it with each of the remaining per-effect rules in turn, in order of highest weight.

Each time (in $CombinePrecons$) the process combines the current precondition $v_{rule}$ with the precondition from the next per-effect rule $v_i$, which we will name $v_{next}$, into a candidate precondition $v_{candidate}$. This includes resolving any conflicts between $v_{rule}$ and $v_{next}$. We now have a candidate precondition $v_{candidate}$ which is a merge of $v_{rule}$ and $v_{next}$. The process refines $v_{candidate}$ further by testing it against a set of alternatives in $SimplifyPrecons$ and setting $v_{candidate}$ to the best result. Now $v_{candidate}$ is tested against the original $v_{rule}$, using $AcceptPrecons$. If $v_{candidate}$ is accepted, $v_{rule}$ is updated to $v_{candidate}$. Similarly, $e_{next}$ is tested against the original set of effects $e_{rule}$, using $AcceptEffects$. If $e_{next}$ is accepted, $e_{rule}$ is updated to $e_{rule} \cup e_{next}$. Finally, the process refines $e_{rule}$ by testing it against a set of alternatives in $SimplifyEffects$ and setting $e_{rule}$ to the best result. In the next section we describe each of the subprocedures in detail.

## 6.2 ALGORITHM DETAILS

*CombinePrecons*: In attempting to combine $(v_{rule}, e_{rule})$ with $(v_{next}, e_{next})$, the first check is whether $e_{next}$ contradicts any effect in $e_{rule}$. Effects conflict if both rules predict change to a fluent, but the rules have different values for the fluent in their preconditions. (For example, the two per-effect rules for (ontable arg$_1$) in Figure 2 conflict.) If there is a conflict, the new rule is rejected, as we assume only one rule per action, and the higher weighted baseline rule is more likely to be correct.

Second, *CombinePrecons* combines the preconditions on every bit which is not locked (listed in $locks$) and does not conflict, i.e., $\forall i\ v_{rule,i} = v_{next,i}$ or $v_{rule,i} = *$:

$$v_{candidate,i} = \begin{cases} v_{rule,i}, & \text{if } v_{rule,i} = v_{next,i} \text{ or } i \in locks \\ v_{rule,i}, & \text{if } v_{next,i} = * \\ v_{next,i}, & \text{if } v_{rule,i} = * \text{ and } i \notin locks. \end{cases}$$

For conflicts, where $v_{rule,i} \neq v_{next,i}$ and $v_{rule,i}, v_{next,i} \in \{1, -1\}$, *CombinePrecons* decides which value each conflict bit should take in $v_{candidate}$, as follows.

For each conflicting fluent, there are three possible values the fluent could take in the preconditions of the true rule: $*$ (unobserved), 1 (true) or $-1$ (false). The weight $weight_e$ (for each effect $e$ in $e_{rule}$) of each variant is calculated (with the values of other conflicting fluents set to $*$). The preferred variant is where the value is $*$, indicating a non-discriminative feature, and giving the simplest precondition. However, a variant is only acceptable if the weight of the resulting precondition is positive for all effects in $e_{rule}$, since then the new precondition still predicts the same effects as the current precondition $v_{rule}$. If accepted, the fluent is locked at the $*$ value, to prevent later, possibly noisy rules, from resetting it. Locked fluents are recorded in the $locks$ variable (see Figure 3). If the $*$-variant is unacceptable, then the (1)-valued or ($-1$)-valued cases are considered, provided they have positive weights on all the effects. If both variants are acceptable, whichever has the highest average weight over all the effects is selected. If neither variant is acceptable then the conflict is unresolved for this fluent. As long as the conflicts on every fluent are resolved, the rule combination process can continue with the new candidate precondition. If not, the current rule is rejected (and *CombinePrecons* returns $v_{rule}$).

*SimplifyPrecons*: Once *CombinePrecons* has generated a candidate precondition $v_{candidate}$, *SimplifyPrecons* considers alternative, less specific preconditions. It creates a set of alternatives $v_{candidate \setminus i}$ for each bit $i$ in $v_{candidate}$ which differs from $v_{rule}$. $v_{candidate \setminus i} = v_{candidate}$ except at bit $i$ where $v_{candidate \setminus i} = *$. Whenever the filtering function *AcceptPrecons* rates $v_{candidate}$ as worse than any $v_{candidate \setminus i}$, the associated fluent $v_{candidate,i}$ is set to $*$.

*SimplifyEffects*: In light of the new preconditions, *SimplifyEffects* tests if any of the effects should be removed from the new $v_{rule}$. For instance, more specific preconditions may lower the incidence of some effects (as seen in the training data) to the extent that *AcceptEffects* rejects them. Each effect is tested against all the other effects by *AcceptEffects* and, if rejected, removed from $e_{rule}$.

*AcceptPrecons*: In the precondition filtering function, before even making a comparison between preconditions, the new precondition $v_{new}$ must be checked to ensure that it is consistent with the classifiers and supported by the training data. For this we require a notion of coverage of the training set. Coverage is defined to account for partial observability, so that precondition $v_{new}$ covers example $x$ at effect $e$ (denoted $covers_e(v_{new}, x)$) if none of the fluents in the example state contradict the fluents in the rule preconditions, and $e$ is in both the example state changes and the rule effects. Now $v_{new}$ can form a rule precondition if:

1. $v_{new}$ is consistent with the classifiers: for each $e \in e_{rule}$, $v_{new}$ should be classified by the classifier $C_{a,e}$ as predicting change, that is,
$\forall e \in e_{rule}\ weight_e(v_{new}) > 0$; and

2. $v_{new}$ is supported by the training data: for each $e \in e_{rule}$, $v_{new}$ should cover at least one training example where $e$ changed, that is,
$\forall e \in e_{rule}\ |\{x : covers_e(v_{new}, x)\}| > 0$.

Both should be considered, as weight alone may permit rules which do not cover any training examples, while coverage alone may allow negatively weighted rules.

Additionally, *AcceptPrecons* uses differences in precision and recall to identify and reject any new precondition which performs significantly worse than the existing precondition. It rejects preconditions where either the precision or recall on the training set drops substantially for any $e \in e_{rule}$. Since precision and recall is a trade-off, the comparison is made using the F-score[2] for precondition $pre$ at effect $e$: $F_{pre,e}$. Ideally we want the new precondition to improve on (or at least not worsen) the F-score, but we must introduce some tolerance to account for the effects of noise.

For instance, suppose $v_{new} = \langle 1, 1, *, * \rangle$ is more general than $v_{rule} = \langle 1, 1, 0, * \rangle$, and that $v_{new}$ is in fact the true rule. The F-score for $v_{rule}$ is calculated on the subset of training examples which $v_{rule}$ covers, while the F-score for $v_{new}$ also includes training examples which $\langle 1, 1, 1, * \rangle$ covers. If the training set happens to have a higher proportion of training examples with a noisy outcome covered by $\langle 1, 1, 1, * \rangle$ than $\langle 1, 1, 0, * \rangle$ then the F-score for $v_{new}$ can be lower than for $v_{rule}$. To account for such effects of noise, we allow the new F-score $F_{v_{new},e}$ to drop to some fraction $\epsilon_p$ of the F-score for the existing precondition $F_{v_{rule},e}$, for any effect $e \in e_{rule}$. For new F-scores below this value, the new precondition is rejected.

---

[2]F-score is the harmonic mean of precision and recall (true positives/predicted changes and true positives/actual changes, respectively) (Van Rijsbergen, 1979).

*AcceptEffect*: The effects filtering function similarly compares F-scores. Given $(v_{rule}, e_{rule})$ it compares how well $v_{rule}$ predicts a new effect $e_{new}$ relative to how well it predicts each $e \in e_{rule}$: specifically it compares $F_{v_{rule},e_{new}}$ to $F_{v_{rule},e}$ for each $e \in e_{rule}$. This identifies effects which are inconsistent with the other effects in terms of precision and recall. In particular, effects which occur in far fewer examples than other effects are identified in this way: these are likely to be caused by noise, or could be conditional effects. An effect is rejected by the function if its F-score is less than some fraction $\epsilon_e$ times the F-score on any other effect of the same rule.

In our evaluation, $\epsilon_p$ and $\epsilon_e$ were set to 0.95 and 0.5 respectively, and not varied across the domains. The values were selected empirically via experiments on a holdout dataset from one experimental domain (ZenoTravel).

### 6.3 RULE COMBINATION EXAMPLE

We now consider an example of one iteration of the rule combination process. Working with the BlocksWorld domain `stack` action, suppose the current rule is $(\langle *, -1, -1, 1, -1, 1, 1, *, * \rangle, \{1, 3, 5\})$ corresponding to the precondition (AND (NOT(clear arg$_1$)) (NOT(ontable arg$_1$)) (holding arg$_1$) (NOT(on arg$_1$ arg$_2$)) (clear arg$_2$) (ontable arg$_2$)) and effects {(clear arg$_1$) (holding arg$_1$) (clear arg$_2$)}. We try to combine this with the new rule $(\langle *, *, -1, *, -1, *, -1, *, 1 \rangle, \{2\})$ corresponding to the precondition (AND (NOT(ontable arg$_1$)) (NOT(on arg$_1$ arg$_2$)) (NOT(ontable arg$_2$)) (on arg$_2$ arg$_1$)) and effects {(ontable arg$_1$)}.

*CombinePrecons* finds no conflicts in the effects, and generates the candidate precondition $\langle *, -1, -1, 1, -1, 1, ?, *, 1 \rangle$ where ? denotes a conflicting fluent. To resolve the conflict the weights of the vectors $\langle *, -1, -1, 1, -1, 1, *, *, 1 \rangle$, $\langle *, -1, -1, 1, -1, 1, 1, *, 1 \rangle$ and $\langle *, -1, -1, 1, -1, 1, -1, *, 1 \rangle$ are calculated for each of the $C_{stack,(clear\ arg_1)}$, $C_{stack,(holding\ arg_1)}$ and $C_{stack,(clear\ arg_2)}$ classifiers. If $\langle *, -1, -1, 1, -1, 1, *, *, 1 \rangle$ is accepted as $v_{candidate}$, *SimplifyPrecons* would consider the alternative precondition $\langle *, -1, -1, 1, -1, 1, *, *, * \rangle$ as the last bit in $v_{candidate}$ was different in $v_{rule}$.

Assuming $\langle *, -1, -1, 1, -1, 1, *, *, * \rangle$ is accepted as $v_{candidate}$ by *SimplifyPrecons*, it is compared to the original $v_{rule}$ (the only difference now is that the bit corresponding to (ontable arg$_2$) is unset in $v_{candidate}$). $v_{candidate}$ has higher weight and so *AcceptPrecons* accepts $v_{candidate}$ and $v_{rule} := v_{candidate}$. Conversely the new effect is rejected by *AcceptEffects* so $e_{rule} = \{1, 3, 5\}$ as before. Finally, *SimplifyEffects* uses *AcceptEffects* to test the relative prediction performance of the new $v_{rule}$ on each effect, with no changes. The new rule is $(\langle *, -1, -1, 1, -1, 1, *, *, * \rangle, \{1, 3, 5\})$.

## 7 EXPERIMENTS

We tested our approach on several simulated domains taken from the International Planning Competition (IPC) at http://ipc.icaps-conference.org/. The domains differ in terms of the number and arity of actions and predicates, and the number and hierarchy of types. The main domain characteristics are detailed in Table 1.

Sequences of random actions and resulting states were generated from the PDDL domain descriptions and used as training and testing data. All data was generated using the Random Action Generator 0.5 available at http://magma.cs.uiuc.edu/filter/, modified to also generate action failures. Table 2 shows the numbers of objects used in training and testing data for each domain.

Ten different randomly generated training and testing sets were used. Training and testing sets were sequences of 20,000 and 2,000 actions respectively. Both sequences contained an equal mixture of successful and unsuccessful actions (where some precondition of the action was not satisfied, and so no change occurred in the world). In some domains (e.g. Rovers), portions of the state space can only be traversed once, and in these cases multiple shorter sequences of 400 actions were generated from randomly generated starting states. In line with previous work (Amir and Chang, 2008), incomplete observations were simulated by randomly selecting a fraction (10%, 25% or 50%) of fluents (including negations) from the world to observe after each action. The remaining fluents were discarded and the reduced state vector was generated from the observed fluents. Sensor noise was simulated similarly by flipping the value of each bit in the state vector with probability 1% and 5%.

### 7.1 RESULTS

We first tested the performance of different kernels on learning the implicit action model, comparing results for a standard (non-kernelised) perceptron, a voted (non-kernelised) perceptron, and a voted kernel perceptron. Both the DNF kernel and k-DNF kernel with $k = 2, 3$ and $5$ were tested. Performance was measured in terms of the F-score of the predictions on the test sets.

The fully observable, noiseless cases are easily learnt by any of the perceptrons tested. After 5,000 training examples, the F-score on the test set is 1, in almost all cases. Per-formance of the voted perceptron, with or without the various kernels, is almost identical (results not shown). With the introduction of unobserved fluents or noise, the voted perceptron performs better than the standard perceptron. However, the DNF kernel does not improve performance, with the unkernelised voted perceptron learning significantly more accurate action models. In contrast, the k-DNF kernels all produce significantly more accurate models than the DNF kernel or no kernel ($p < 0.05$, repeated measures ANOVA with post-hoc Bonferroni t-test). Figure 4 gives a comparison of the relative performance of each model. The use of k-DNF kernels therefore represents a significant improvement on previous work which used only the DNF kernel. In light of these results, the 3-DNF kernel was selected for the remainder of the experiments.

Table 1: Domain Characteristics

| Domain | Actions | | Predicates | |
|---|---|---|---|---|
| | No. | Max arity | No.(+types) | Max arity |
| BlocksWorld | 4 | 2 | 5 | 2 |
| Depots | 5 | 4 | 6 (+6) | 2 |
| ZenoTravel | 5 | 6 | 8 (+4) | 2 |
| DriverLog | 6 | 4 | 6 (+4) | 2 |
| Rovers | 9 | 6 | 25 (+7) | 3 |

Table 2: Number of Objects in Training and Testing Worlds

| Domain | Training | Testing |
|---|---|---|
| BlocksWorld | 13 blocks | 30 blocks |
| Depots | 1 depot | 4 depots |
|  | 2 distributors | 4 distributors |
|  | 2 trucks | 4 trucks |
|  | 3 pallets | 10 pallets |
|  | 3 hoists | 8 hoists |
|  | 10 crates | 8 crates |
| ZenoTravel | 5 cities | 10 cities |
|  | 3 planes | 5 planes |
|  | 7 people | 10 people |
| DriverLog | 3 road junctions | 20 road junctions |
|  | 3 drivers | 5 drivers |
|  | 7 packages | 25 packages |
|  | 3 trucks | 5 trucks |
| Rovers | 2 rovers | 4 rovers |
|  | 4 waypoints | 8 waypoints |
|  | 3 objectives | 4 objectives |
|  | 3 cameras | 4 cameras |
|  | 3 modes | 3 modes |
|  | 2 stores | 4 stores |
|  | 1 lander | 1 lander |

Next, we extracted explicit rules from the implicit action models. There was no statistically significant difference between the F-scores of predictions made by the perceptron models and those made by the extracted rules (repeated measures ANOVA, $p > 0.05$). We also compared the resulting models to the original domain descriptions using a measure of error rate (Zhuo *et al.*, 2010). The error rate for a single action is defined as the number of extra or missing fluents in the preconditions and effects ($E_{pre}$ and $E_{eff}$ respectively) divided by the number of possible fluents in the preconditions and effects ($T$): $Error(a) = \frac{1}{2T}(E_{pre} + E_{eff})$. The error rate of a domain model with a set of actions $A$ is: $Error(A) = \frac{1}{|A|} \sum_{a \in A} Error(a)$.

The error rates indicate that the learnt models are close to the actual STRIPS domain definitions, falling below 0.1 after around 5,000 examples in all cases (Figure 5). In particular, for fully observable, noiseless domains the correct STRIPS model is given by the extracted rules in fewer than 2,000 training examples, except for the most complex domain, Rovers. Comparisons with other approaches in the literature are difficult due to differences in the learning settings. Nevertheless it is notable that the error rates of the learnt action models are low in comparison to action models learnt by (Yang *et al.*, 2007) for the same domains: their error rates at 90% observability (the highest reported) range from around 0.04 (ZenoTravel) to 0.1 or above (DriverLog and Depots) to more than 0.6 (Rovers). An example action model is shown in Figure 6, demonstrating that the method derives compact STRIPS-like rules even with high levels of incompleteness and noise in the observations.

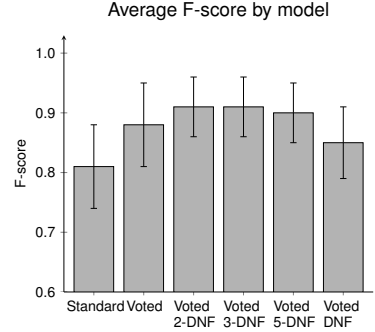

Figure 4: Comparison of the performance of different perceptrons learning action models from 20,000 random actions in STRIPS domains, averaged across all domains, levels of noise and partial observability. Error bars are standard error. Performance is significantly different between models which use a k-DNF kernel and those which do not.

We also calculated the F-scores for predictions made by the learnt rules on our (noiseless, fully observable) test sets (Figure 5). The F-scores are above 0.9 for all noise levels at 25% observability and above, for all domains except Rovers, indicating that in practice the rules correctly predict most fluents. The Rovers F-scores are somewhat lower, because there are fewer training examples per action than for the other domains, and more possible fluents.

Furthermore, our learning is fast. The longest-running example in the experiments (Rovers with 20,000 training examples, 5% noise, fully observable) takes under 1.5 hours on a single Intel Xeon 5160 processor to train the classifiers, and run rule extraction and combination. The ZenoTravel example (Figure 6) runs in under 2 minutes.

## 8 CONCLUSIONS AND FUTURE WORK

The results demonstrate that our approach successfully learns STRIPS operators from noisy, incomplete observations, in contrast to previous work which either generates explicit operators but cannot tolerate noise and incomplete examples, or tolerates noise and incomplete examples but does not generate explicit operators. We also show empirically that the 3-DNF kernel is a more appropriate choice

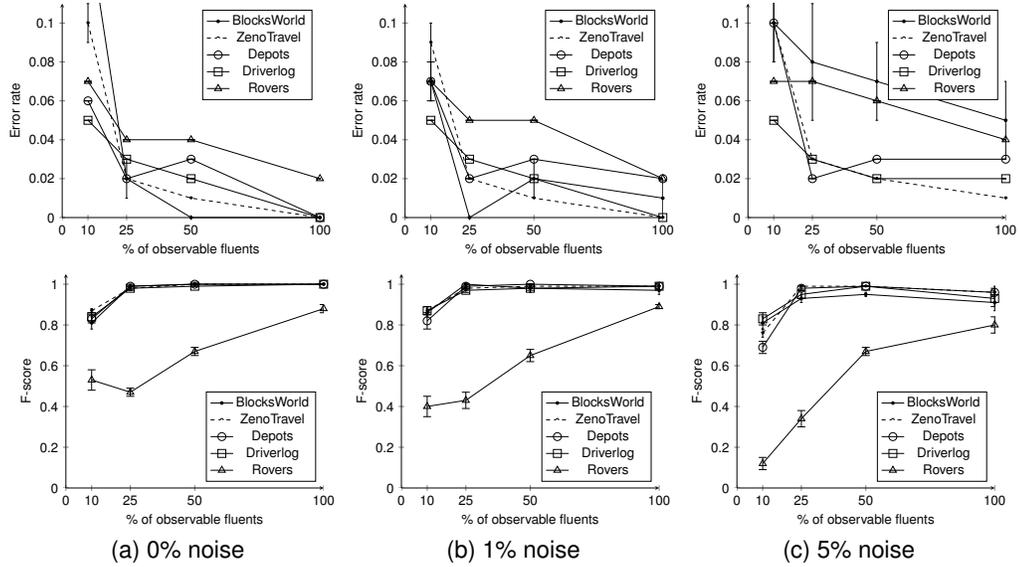

(a) 0% noise  (b) 1% noise  (c) 5% noise

Figure 5: Results from learning explicit action rules from 5,000 training examples at varying levels of observability and noise in simulated planning domains. The error rate measures errors in the learnt domain model relative to the actual domain model (above). The F-score measures performance of the rules on fully observable, noiseless test domains (below).

than the DNF kernel for learning in this setting.

Our approach depends on decomposing the learning problem into two stages: learning implicit action models and then deriving explicit rules from the implicit models. Crucially, the implicit models produce noise-free, complete observations *for the domain model which has been learnt*. An alternative approach to our rule derivation process would be to apply existing action model learning techniques to the observations produced by the implicit models. However such an approach effectively restarts the learning process, ignoring information already learnt and available in the perceptron models, and so is likely to be less efficient.

Our approach also depends on the STRIPS scope assumption (SSA) which essentially identifies the objects which are relevant to the action and fixes their roles. In real-world scenarios the SSA may not apply. Without the SSA, during learning we must also consider state relating to objects which are not listed in the action parameters. Implicit action models in this setting may be learnt using a graphical representation of states combined with a suitable graph kernel (Mourão, 2012). In future work we therefore plan to extend our rule extraction method to derive rules from classifiers trained with graphical state representations. Additional steps will be required to efficiently handle the complexity introduced by the requirement to perform comparisons between graphical state descriptions. There are positive results in PAC-learning existential conjunctive and k-DNF concepts in noise-free structural domains with Boolean relations (Haussler, 1989; Valiant, 1985), which apply to learning from the implicit models, suggesting that our approach will scale to graphical state representations.

```
(:action DEBARK
:parameters (?x1 ?x2 ?x3 )
:precondition (AND (in ?x1 ?x2) (at ?x2 ?x3))
:effect (AND (at ?x1 ?x3) (NOT(in ?x1 ?x2))))

(:action BOARD
:parameters (?x1 ?x2 ?x3 )
:precondition (AND (at ?x1 ?x3) (at ?x2 ?x3))
:effect (AND (NOT(at ?x1 ?x3)) (in ?x1 ?x2)))

(:action FLY
:parameters (?x1 ?x2 ?x3 ?x4 ?x5 )
:precondition (AND (at ?x1 ?x2) (fuel-level ?x1 ?x4)
        (next ?x5 ?x4))
:effect (AND (NOT(at ?x1 ?x2)) (at ?x1 ?x3)
        (NOT(fuel-level ?x1 ?x4)) (fuel-level ?x1 ?x5)))

(:action ZOOM
:parameters (?x1 ?x2 ?x3 ?x4 ?x5 ?x6 )
:precondition (AND (at ?x1 ?x2) (fuel-level ?x1 ?x4)
        (next ?x6 ?x5) (next ?x5 ?x4))
:effect (AND (NOT(at ?x1 ?x2)) (at ?x1 ?x3)
        (NOT(fuel-level ?x1 ?x4)) (fuel-level ?x1 ?x6)))

(:action REFUEL
:parameters (?x1 ?x2 ?x3 ?x4 )
:precondition (AND (fuel-level ?x1 ?x3) (next ?x4 ?x3)
        (next ?x3 ?x4) (at ?x1 ?x2))
:effect (AND (NOT(fuel-level ?x1 ?x3)) (fuel-level ?x1 ?x4)))
```

Figure 6: Explicit action model output for the ZenoTravel domain after 5,000 training examples with 10% observability and 5% noise. Missing fluents are in bold italic, incorrect fluents in bold. The error rate of this example is 0.05. Such imperfect rules have quite small effects on performance (F-score 0.85 in this case), but will in future work be improved by eliminating low reliability classifiers.

## Acknowledgements

The authors are grateful to the reviewers of this and previous versions of this paper for helpful comments. This work was partially funded by the European Commission through the EU Cognitive Systems project Xperience (FP7-ICT-270273) and the UK EPSRC/MRC through the Neuroinformatics and Computational Neuroscience Doctoral Training Centre, University of Edinburgh.